\documentclass[letterpaper, 10 pt, conference]{ieeeconf}

\IEEEoverridecommandlockouts                              

\usepackage{graphicx}
\usepackage{multicol}
\usepackage[hidelinks,breaklinks]{hyperref}
\usepackage{bm}
\usepackage{url}
\usepackage{graphicx} 
\usepackage{mathptmx} 

\usepackage{subcaption}
\usepackage[mathscr]{euscript}  
\usepackage{amsmath} 
\usepackage{amssymb}  
\usepackage{flushend}
\usepackage{multirow}
\usepackage{siunitx}
\usepackage{xcolor}
\usepackage{adjustbox}
\usepackage{array}
\usepackage{multirow}
\usepackage{booktabs}
\let\labelindent\relax
\usepackage[inline]{enumitem}
\usepackage[hang,flushmargin]{footmisc}
\usepackage{microtype}
\usepackage{amsmath}
\usepackage{threeparttable}
\usepackage{caption}
\usepackage[resetlabels]{multibib}
\usepackage{cite}

\newcites{New}{References}

\renewcommand{\eqref}[1]{Eq.~(\ref{#1})}

\newcommand{\nset}{\set}

\newcolumntype{P}[1]{>{\centering\arraybackslash}p{#1}}

\newcommand{\metricname}{PDC-Q }
\newcommand{\netname}{PDcast }

\usepackage{pifont}
\renewcommand{\baselinestretch}{0.98}

\usepackage{color,array}

\makeatletter
\def\zapcolorreset{\let\reset@color\relax\ignorespaces}
\def\colorrows#1{\noalign{\aftergroup\zapcolorreset#1}\ignorespaces}
\makeatother

\newcommand{\set}[1]{\left\{#1\right\}}

\title{\LARGE \bf 
Panoptic-Depth Forecasting
}

\author{Juana Valeria Hurtado$^*$, Riya Mohan$^*$, Abhinav Valada
\thanks{$^*$These authors contributed equally.}
\thanks{Department of Computer Science, University of Freiburg, Germany.}%
\thanks{This work was funded by the German Research Foundation (DFG) Emmy Noether Program grant number 468878300.}}

\begin{document}

\maketitle
\thispagestyle{empty}
\pagestyle{empty}

\raggedbottom

\begin{abstract}
Forecasting the semantics and 3D structure of scenes is essential for robots to navigate and plan actions safely. Recent methods have explored semantic and panoptic scene forecasting; however, they do not consider the geometry of the scene. In this work, we propose the panoptic-depth forecasting task for jointly predicting the panoptic segmentation and depth maps of unobserved future frames, from monocular camera images. To facilitate this work, we extend the popular \mbox{KITTI-360} and Cityscapes benchmarks by computing depth maps from LiDAR point clouds and leveraging sequential labeled data. We also introduce a suitable evaluation metric that quantifies both the panoptic quality and depth estimation accuracy of forecasts in a coherent manner. Furthermore, we present two baselines and propose the novel \netname architecture that learns rich spatio-temporal representations by incorporating a transformer-based encoder, a forecasting module, and task-specific decoders to predict future panoptic-depth outputs. Extensive evaluations demonstrate the effectiveness of \netname across two datasets and three forecasting tasks, consistently addressing the primary challenges. We make the code publicly available at \url{https://pdcast.cs.uni-freiburg.de}
\end{abstract}

\section{Introduction}

The ability to predict the semantics and depth map of the scene is crucial for enabling robots to operate effectively in real-world environments~\cite{gosala2023skyeye,mohan2022perceiving,mohan2024panoptic}. Furthermore, forecasting the future semantics and spatial 3D scene structure is vital for robots to perform safe interaction and planning. For example, in the context of navigation and autonomous driving, it is essential to forecast the future identities and locations of all the elements of the scene, such as vehicles, roads, and obstacles, for intelligent decision-making. However, estimating both the semantics and geometry of future frames is a challenging problem given the complex scene dynamics, the exponential space-time dimensionality, and the non-deterministic nature of the future.\looseness=-1

Recent advances in scene forecasting have made significant progress in predicting future perceptions of individual tasks~\cite{boulahbal2022forecasting,graber2021panoptic,vsaric2021dense,graber2022joint}. One such task is panoptic segmentation forecasting, which predicts pixel-level semantics and instance IDs of unobserved future camera frames~\cite{graber2021panoptic, graber2022joint}. While this offers a rich semantic understanding of scene evolution, it lacks crucial geometric information, which is essential for planning safe actions and disambiguating perceptual aliasing. Conversely, depth forecasting offers geometric insights into the future scene by estimating the relative distances~\cite{boulahbal2022forecasting}, but it typically does not consider semantic information. Jointly predicting the semantics and depth from a single image has been shown to benefit both tasks by leveraging complementary cues and inductive transfer~\cite{qiao2021vip, he2023towards}. We aim to further exploit these advantages in scene forecasting by jointly predicting the spatial 3D panoptic structure of the evolving scene. 

This task introduces several challenges. Traditional methods that employ specialized networks for each task often achieve strong performance but at the cost of increased computational complexity, as multiple models need to be trained and deployed in parallel. Conversely, learning shared features that capture both panoptic segmentation and depth information in a unified framework is more computationally efficient but significantly more difficult. This is because these tasks require different spatial, semantic, and structural reasoning. Predicting future scenes adds another layer of complexity, as the model must account for the temporal evolution of objects and their relationships in 3D space. This includes challenges such as anticipating the movement and interaction of dynamic elements and handling changes in lighting and appearance. Capturing both the semantic structure and depth of objects as they evolve over time requires not only accurate feature extraction but also robust temporal modeling, making this joint forecasting task particularly challenging.

\begin{figure}
    \centering
    \includegraphics[width=0.95\linewidth]{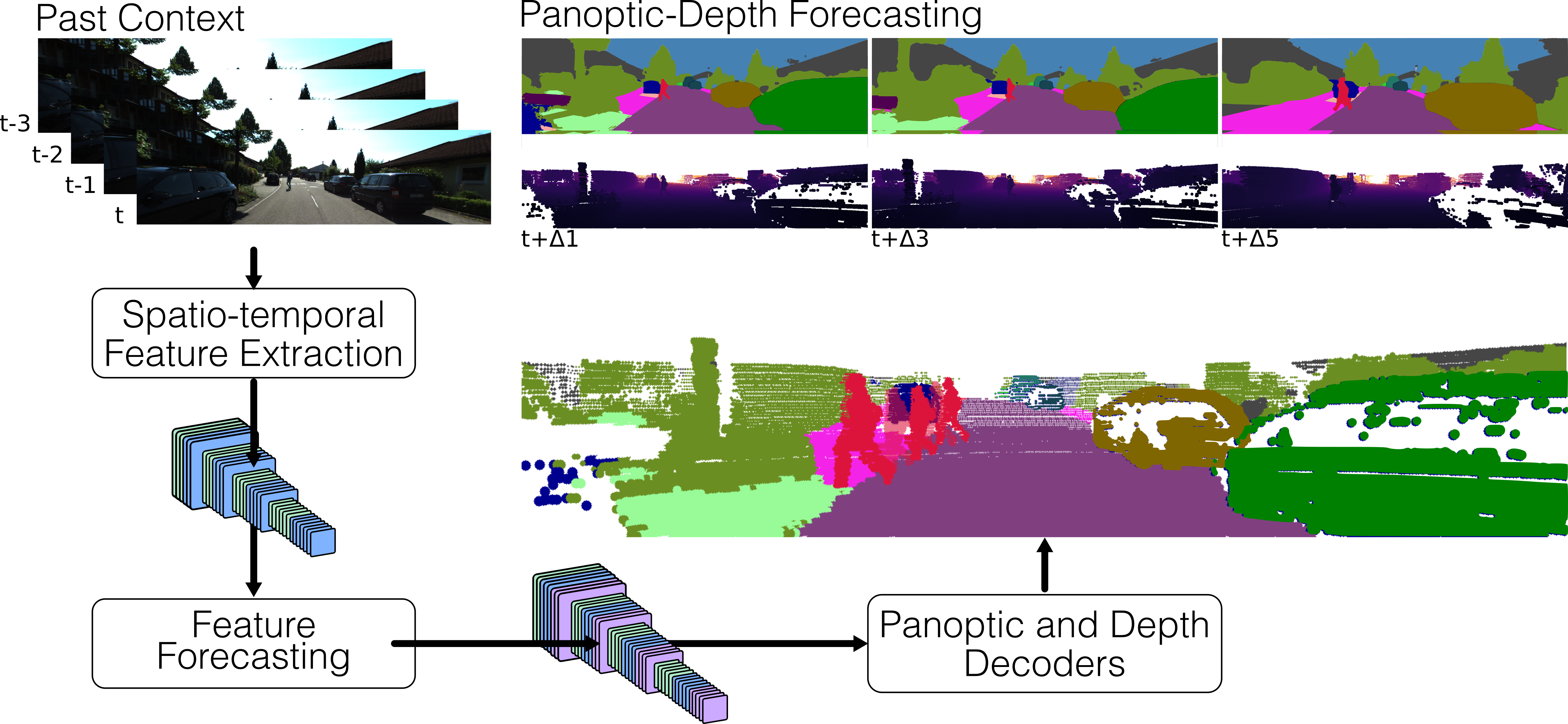}
    \caption{Panoptic-depth forecasting learns rich spatio-temporal representations to jointly predict the pixel-level semantic category, instance ID, and depth value of unobserved future frames.}
    \label{fig:teaser}
    \vspace{-0.3cm}
\end{figure}

In this work, we introduce panoptic-depth forecasting, a novel perception task that forecasts the semantic categories, instance IDs, and depth values of the scene from a sequence of past monocular camera images. To the best of our knowledge, this is the first work to forecast panoptic and depth predictions simultaneously. In addition to the task definition, we establish a benchmark using two standard datasets, KITTI-360~\cite{liao2022kitti} and Cityscapes~\cite{cordts2016cityscapes}, containing panoptic segmentation labels and depth maps. We introduce two baselines by combining state-of-the-art panoptic segmentation forecasting and depth forecasting methods. Furthermore, we propose the \metricname metric that coherently quantifies the performance of the panoptic-depth forecasting task by incorporating panoptic quality and depth accuracy of different numbers of frames in the future. As the first novel approach to address this task, we propose the \netname architecture that consists of a transformer-based encoder, a multi-scale spatio-temporal aggregation module, and task-specific decoders for predicting panoptic-depth forecasts. We perform extensive evaluations on the panoptic-depth forecasting task, as well as panoptic forecasting and depth forecasting tasks individually, to show the benefits of joint learning. By comparing results across two datasets and three forecasting tasks, we demonstrate the effectiveness of our proposed approach.

We summarize our main contributions as follows:
\begin{enumerate}[topsep=0pt]
    \item The panoptic-depth forecasting task for simultaneously predicting future panoptic segmentation and depth maps from camera images. We formulate the task and identify its challenges.
    \item The novel \netname architecture that effectively addresses panoptic-depth forecasting by learning rich spatio-temporal representations.
    \item The \metricname metric for coherently quantifying the performance of panoptic-depth forecasting methods.
    \item Two novel baseline methods by combining state-of-the-art panoptic forecasting and depth forecasting methods.
    \item Extensive experiments and ablation study on two challenging datasets.
    \item We make our code available at \url{https://pdcast.cs.uni-freiburg.de}.

\end{enumerate}

\section{Related Work}
In this section, we review related work in general scene forecasting, panoptic segmentation forecasting, and depth forecasting.

{\parskip=2pt
\noindent\textit{Scene Forecasting} has largely focused on object trajectory prediction, utilizing deterministic and probabilistic models and temporal feature learning modules such as RNNs~\cite{zhang2019sr}, normalizing flows~\cite{rhinehart2019precog}, and transformers~\cite{yuan2021agentformer}. However, these approaches overlook the broader scene context, such as spatial distribution and categories of the rest of the scene elements, which are crucial for comprehensive scene understanding and decision-making. A more comprehensive future scene prediction was proposed through camera image forecasting where semantic and instance maps along with optical flow are used to synthesize future camera frames~\cite{wu2020future}. However, experiments have demonstrated that forecasting segmentation maps yields better results than forecasting raw camera frames and then segmenting them~\cite{luc2017predicting}. Following, methods for forecasting semantic segmentation and instance segmentation~\cite{yuan2021agentformer} have been developed independently. Approaches for semantic segmentation forecasting propagate multi-scale features from convolutional encoders, using a convolution decoder~\cite{luc2017predicting} or flow wrapping~\cite{saric2020warp}. For instance segmentation, most approaches forecast the intermediate latent representations of the detected mask using CNNs~\cite{luc2018predicting} and ConvLSTMs~\cite{sun2019predicting}.} 

{\parskip=2pt
\noindent\textit{Panoptic Segmentation Forecasting}: Closer to our proposed task, panoptic segmentation forecasting predicts the semantic category and instance ID of future frames. Graber~\textit{et~al.}~\cite{graber2021panoptic} use a specialized network that incorporates camera images, pre-computed depth maps, and odometry to independently forecast each instance. The remaining 'stuff' or background classes are forecasted by warping the semantics of the input frame to the future using a 3D rigid-body transformation. The final panoptic forecasting output is the heuristic combination of masks and the background. Subsequent work~\cite{graber2022joint} forecasts all instances as foreground using a transformer-based architecture and refines the predictions with depth maps and odometry. Despite the advancements, these methods have two main drawbacks. First, they rely on external depth and odometry data. Second, they lack a geometric understanding of the scene. Our approach addresses these limitations by jointly forecasting panoptic segmentation and depth maps from past raw camera images, providing geometry-aware panoptic forecasts without using additional data.}

{\parskip=2pt
\noindent\textit{Monocular Depth Forecasting} was first addressed as part of RGB-D future synthesis where RGB pixels, semantic maps, and depth maps are forecasted to the future adjacent frame~\cite{qi20193d}. Subsequent work forecasts only semantics and depth maps using a probabilistic generative model~\cite{hu2020probabilistic}. Nag~\textit{et~al.} formulate depth forecasting as a view-synthesis problem, where depth estimation is an auxiliary task of a self-supervised framework that synthesizes views based on learned pose~\cite{nag2022far}. Boulahbal~\textit{et~al.} combine convolutional and transformer modules to generate rich spatio-temporal representations for depth forecasting~\cite{boulahbal2022forecasting}. A recent method learns future depth predictions to improve current depth estimation by iteratively predicting multi-frame features one step ahead~\cite{yasarla2024futuredepth}. In our work, we extend this line of research by jointly forecasting pixel-level semantics, instance IDs, and depth maps, which is crucial for intelligent decision-making.}
\begin{figure*}
    \centering
    \includegraphics[width=0.85\linewidth]{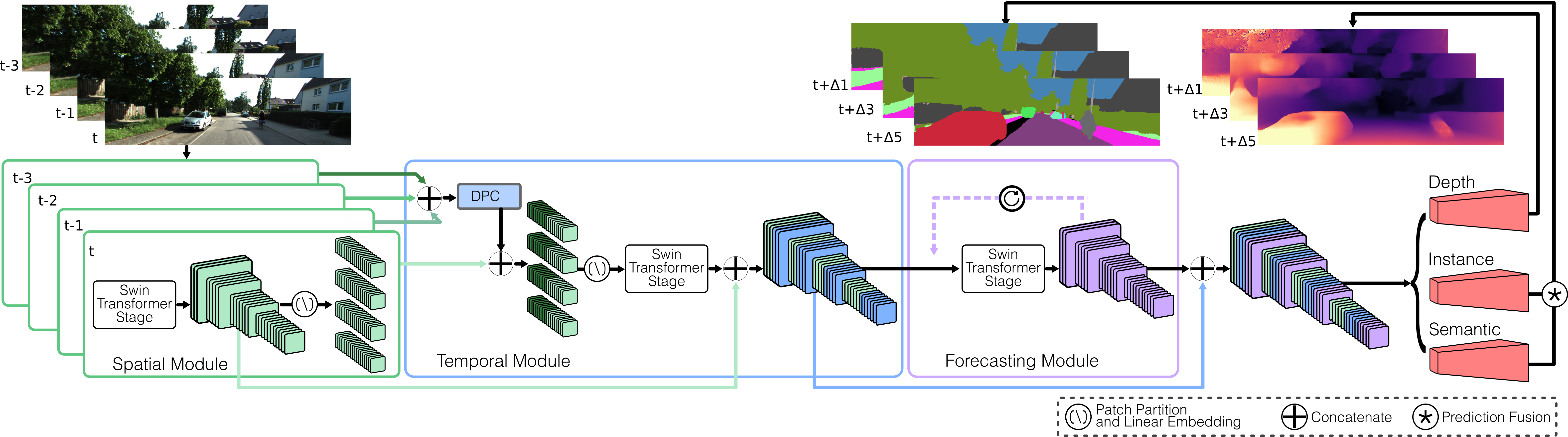}
    \caption{Overview of our proposed \netname architecture for panoptic-depth forecasting. A single transformed-based encoder extracts rich spatio-temporal features from past monocular camera images. The forecasting module then learns to forecast features into the future, which serve as the input to the two decoders for panoptic segmentation and depth estimation.}
    \label{fig:architecture}
    \vspace{-0.3cm}
\end{figure*}

\section{Panoptic-Depth Forecasting}
\subsection{Task Definition}

Given a sequence of observed past camera images $I_{t-k:t} \in \mathbf{R}^{w \times h \times c}$, the goal of panoptic-depth forecasting is to predict a tuple $ (c,id,d)_{t+\Delta}$ for each pixel in unobserved future frames. This tuple forecasts the semantic class, instance ID, and depth value at a future time step $t+\Delta$, where $\Delta$ indicates the number of frames ahead.

\subsection{Evaluation Metric}

We propose a unified metric, Panoptic Depth Forecasting Quality (\metricname), to coherently assess both the accuracy of depth prediction and the panoptic segmentation of future frames at time $t+\Delta$. This metric is adapted from the depth-aware video panoptic segmentation metric, which evaluates panoptic-depth prediction of future frames. Given the predicted object segments $P$
and their ground truth counterparts $G$ for each class $c$, we define \metricname based on the Panoptic Quality (PQ) metric~\cite{hurtado2022semantic} as follows:
\begin{align}
    \text{PDC-Q}_{t+\Delta}^{\lambda}= \frac{1}{|C|}\sum_{c \in C}\frac{\sum_{(p,g)_\in TP_c} IoU(p,g)}{|TP_c|+\frac{1}{}|FP_c|+\frac{1}{2}|FN_c|},
\end{align}
where $TP$, $FN$, and $FP$ are the true positives, false positives, and false negatives that we determine based on the absolute relative depth errors $\delta$ as

{\small
\begin{align}
TP_c &= \{p \in \nset{P} \;|\; \lambda < u \; \& \; IoU(p, g) > 0.5, \; \forall \; g \in \nset{G}\}, \\
FP_c &= \{p \in \nset{P} \;|\; IoU(p, g) <= 0.5, \; \forall \; g \in \nset{G}\}, \\
FN_c &= \{g \in \nset{G} \;|\; IoU(g, p) <= 0.5, \; \forall \; p \in \nset{P}\},
\end{align}}

\noindent{where $u$ is a threshold in $\{0.1,0.25,0.5\}$. \metricname penalizes pixels with large absolute relative depth errors, which are computed based on the depth inlier criteria described in \cite{qiao2021vip}. The final evaluation score is computed by averaging the metric across multiple future predictions at time steps $t+\Delta$. Finally, the overall \metricname is defined as}
\begin{align}
    \text{PDC-Q}^{\lambda} = \sum_{\Delta} \text{PDC-Q}_{t+\Delta}^{\lambda},
\end{align}
providing a coherent evaluation of both depth prediction and panoptic segmentation quality across multiple future frames.

\subsection{Challenges}
Jointly forecasting panoptic segmentation and depth values of future frames presents several significant challenges. This task requires accurate models of both semantic and geometric information. A straightforward approach that uses separate networks for each task would not only increase computational complexity but also fail to exploit complementary cues between the tasks that could enhance performance. An ideal solution for panoptic-depth forecasting involves a unified approach, using a shared backbone to forecast multipurpose features, which enables more efficient and accurate predictions. However, even single-frame models that predict both depth and panoptic segmentation using a shared backbone are scarce. Extending this complexity to future frames introduces additional challenges, particularly given the non-deterministic nature of the future. The complex scene dynamics further increase the difficulty of forecasting their future distributions. Additionally, most current architectures approach scene forecasting deterministically, which may not adequately capture the uncertainties inherent in future prediction.\looseness=-1
 
\subsection{Baselines}
\label{sec:baselines}
To the best of our knowledge, there are no existing methods that tackle panoptic-depth forecasting. Therefore, we propose two baseline models suitable for this task.

\textbf{CODEPS(+)}: The first baseline adapts a convolution-based network that jointly predicts panoptic segmentation and depth estimation from single images. Specifically, we build on a pretrained network that uses ResNet-101 as a shared encoder for depth, semantic, and instance segmentation tasks~\cite{voedisch23codeps}. Similar to \cite{hu2020probabilistic}, we extend this network by extracting features from the encoder's output of the past five frames. These features are then passed through a spatio-temporal convolutional block, which forecasts the features for future frames. The predicted features are subsequently decoded by the pretrained panoptic and depth decoders.

\begin{figure}
  \centering
  {\includegraphics[width=0.9\linewidth]{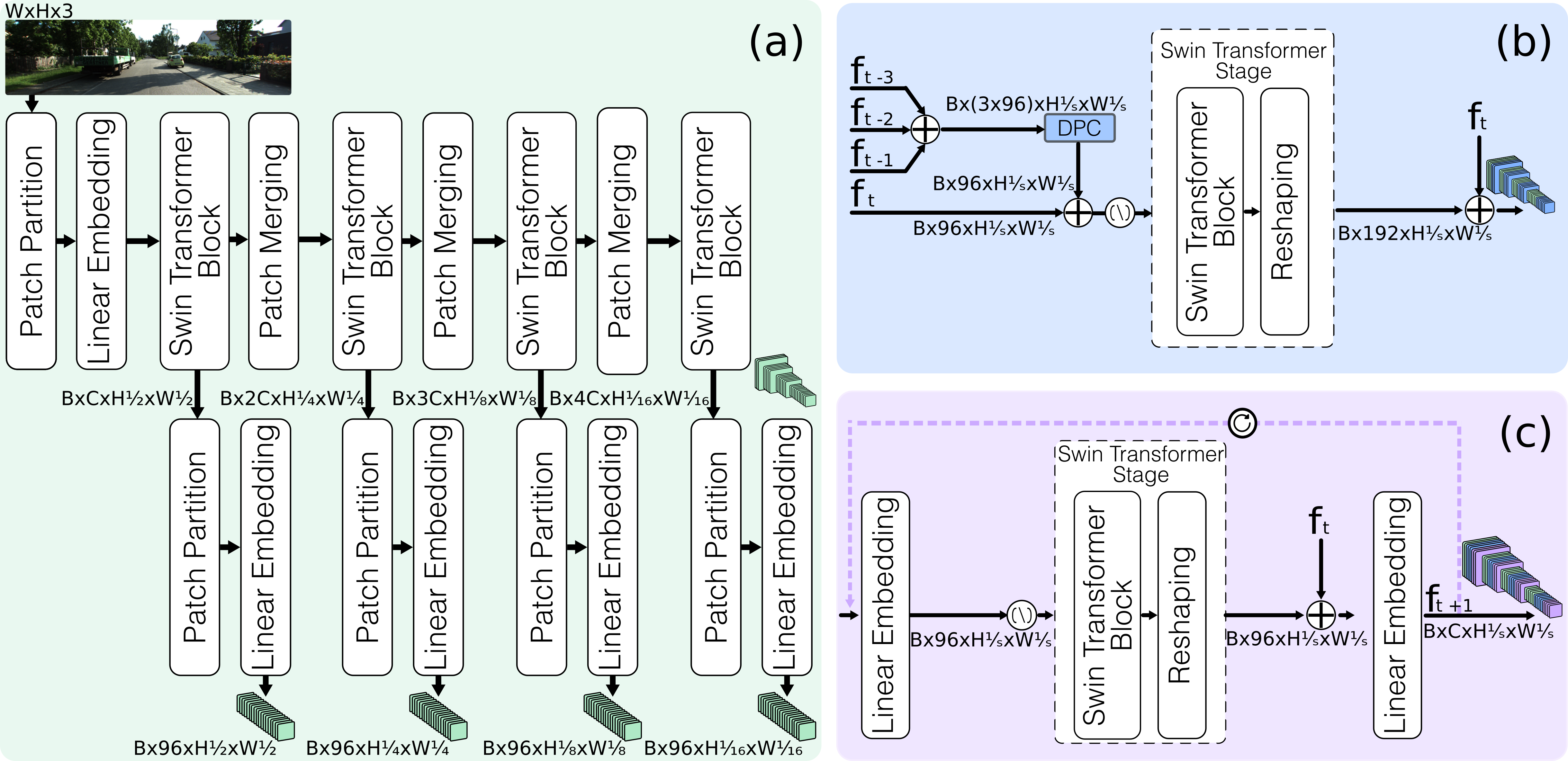}}
\caption{Architecture of the (a)~spatial module, (b)~temporal module, and (c)~forecasting module. The spatial module processes each time frame separately. The spatial and forecasting modules show the process for each scale $s = [2,4,8,16]$.}
\label{fig:spatiotemporal}
\vspace{-0.2cm}
\end{figure}

\textbf{[Depth + PS](+)}: The second baseline is a combination of two separate forecasting models. The first model focusing on depth forecasting employs a Swin Transformer backbone and Monodepth2 depth decoder~\cite{godard2019digging}. The second model, designed for panoptic forecasting, also utilizes a Swin Transformer backbone and fuses the outputs from a semantic head and an instance head~\cite{cheng2020panoptic}. We train both models without sharing weights despite them having the same encoder and forecasting module depicted in Fig.~\ref{fig:spatiotemporal}. Additionally, we initialize the weights of the decoders with pretrained models from \cite{voedisch23codeps}.

\section{PDCast Architecture}
In this section, we detail the \netname architecture consisting of three key components: a spatio-temporal feature extraction block, a forecasting block, and task-specific decoders for depth estimation and panoptic segmentation. The spatio-temporal feature extraction block leverages a Swin Transformer encoder with multi-head attention to capture spatial and temporal information, while the forecasting block predicts future features. Finally, task-specific decoders take these features as input to predict future depth and panoptic segmentation. Fig.~\ref{fig:architecture} illustrates the \netname architecture.

Our framework follows a multi-task learning paradigm, where a single encoder extracts multi-purpose features that are then processed by task-specific decoders. We include an intermediate forecasting module that predicts future panoptic-depth features from previous frames. This is accomplished in three stages: (1) spatio-temporal feature extraction, (2) feature forecasting, and (3) task-specific decoders for depth and panoptic segmentation. The network takes as input a sequence of camera frames $I_{t-k}$ (where $k=4$) and outputs predictions for panoptic segmentation and depth $(c, id, d)_{t+\Delta}$ for future frames $\Delta = {0, 1, 3, 5}$.

\subsection{Spatio-Temporal Feature Extraction}
\label{sec:forecasting-block}
The Spatio-Temporal Feature Extraction module has two stages: spatial and temporal. The spatial module first uses a Swin Transformer encoder to extract multi-scale features from each camera frame independently. Then, a Swin Transformer stage concatenates these spatial features across time to capture temporal dependencies for each scale.

{\parskip=2pt
\noindent\textit{Spatial Feature Extraction}: We extract spatial features from each past frame using the module shown in Fig.~\ref{fig:spatiotemporal}(a). Each input camera frame $I_{t-k}$ is divided into non-overlapping $4 \times 4$ patches, which are linearly embedded into 96-dimensional feature vectors. These patches are passed through the Swin Transformer encoder, which consists of four hierarchical stages. Each stage includes several Swin Transformer blocks and patch merging operations. The Swin encoder has depths of ${2,2,6,2}$, an embedding size of 96, and attention heads of ${3,6,12,24}$. The Swin Transformer block employs window-based multi-head self-attention to extract spatial features. Patch merging layers reduce spatial resolution and increase feature dimensionality, resulting in a hierarchical multi-scale representation of each frame.
{\parskip=2pt
\noindent\textit{Temporal Feature Extraction}: After extracting spatial features from each frame, a Dense Prediction Cell (DPC) module processes earlier frames ($F_{t-1}, F_{t-2}, F_{t-3}$) to capture long-range context as depicted in Fig.~\ref{fig:spatiotemporal}(b). The output of the DPC is concatenated with the most recent feature map $F_{t}$. This combined tensor is then input to a Swin Transformer stage consisting of a Swin Transformer block and a patch merging layer, which extracts temporal features across the sequence. By leveraging multi-head attention, the network learns correlations between spatial features from different frames, attending to relevant regions at various time steps and improving future scene dynamics prediction. 

The spatio-temporal module produces multi-scale feature maps,  $F_t$, that integrate spatial details from individual frames and temporal relationships across the sequence.

\subsection{Panoptic-Depth Feature Forecasting}

The forecasting module predicts future frames $F_{t+1}$ through a recursive process that sequentially processes multi-scale feature maps derived from the preceding Spatio-Temporal (ST) block $F_t$. At each scale, the feature map from $F_t$ is projected to an embedding dimension of 96 and passed through a Swin Transformer block with a depth of 2 and a varying number of attention heads, specifically [3, 6, 12, 24], tailored for different scales from high to low resolution. The transformed output is reshaped back to its original spatial dimensions and concatenated with the corresponding $F_t$ features using skip connections. For each subsequent future frame $F_{t+2}, F_{t+3},...$, the reshaped features are again concatenated with the corresponding $F_t$ features at the same scale. The concatenated output is then linearly projected back to its original channel dimensions to generate the features for $F_{t+1}$. This process repeats recursively, where the features for $F_{t+k-1}$ are used to compute the features for $F_{t+k}$, effectively capturing temporal dynamics and multi-scale spatial dependencies across all future frames.

\begin{table*}
\caption{Panoptic-depth forecasting results on KITTI-360 and Cityscapes-DVPS datasets. We compare the performance of our proposed \netname with two baselines and the oracle for short-term ($\Delta t = 1$ and $\Delta t = 3$ on KITTI-360 and $\Delta t = 1$ on Cityscapes-DVPS) and mid-term ($\Delta t = 5$) predictions. The presented metrics include our proposed \metricname, PQ (panoptic quality), and RMSE (root mean square error). Oracle $\dagger$ means the reference value. Each sequence in Cityscapes DVPS consists of six frames; we use three past frames and a current frame, leaving two future frames corresponding to  $\Delta = [3,5]$. }
\centering
\footnotesize
\begin{tabular}{@{}lp{2.3cm}|P{1.2cm}P{1.2cm}P{1.2cm}|P{1.2cm}P{1.2cm}P{1.2cm}|P{1.2cm}P{1.2cm}P{1.2cm}@{}}
\toprule
& \multirow{2}{*}{Network} & \multicolumn{3}{c|}{Short term $\Delta t = 1$} & \multicolumn{3}{c|}{Short term $\Delta t = 3$} & \multicolumn{3}{c}{Mid term $\Delta t = 5$} \\
\cmidrule{3-11}
&& \metricname $\uparrow$ & PQ $\uparrow$ & RMSE $\downarrow$ & \metricname $\uparrow$ & PQ $\uparrow$ & RMSE $\downarrow$ & \metricname $\uparrow$ & PQ $\uparrow$ & RMSE $\downarrow$ \\
\midrule
\parbox[t]{2mm}{\multirow{4}{*}{\rotatebox[origin=c]{90}{KITTI-360}}} & Oracle $\dagger$ & $-$ & $45.31$ & $3.94$ & $-$ & $45.31$ & $3.94$ & $-$ & $45.3$ & $3.94$ \\
& CODEPS(+)& $36.25$ & $36.78$ & $4.12$ & $30.01$ & $30.63$ & $4.52$ & $25.39$ & $25.8$ & $4.87$ \\
& [Depth + PS](+) & $38.35$ & $39.05$ & $4.11$ & $32.45$ & $33.00$ & $4.49$ & $28.47$ & $28.95$ & $4.81$ \\
\cmidrule{2-11}
& \netname (Ours) & $\mathbf{41.03}$ & $\mathbf{41.76}$ & $\mathbf{4.04}$ & $\mathbf{36.26}$ & $\mathbf{36.86}$ & $\mathbf{4.23}$ & $\mathbf{32.17}$ & $\mathbf{32.7}$ & $\mathbf{4.44}$ \\
\midrule
\parbox[t]{2mm}{\multirow{4}{*}{\rotatebox[origin=c]{90}{Cityscapes}}} & Oracle $\dagger$ & $-$ & $-$ & $-$ & $-$ & $57.1$ & $3.1$ & $-$ & $57.1$ & $3.1$ \\
& CODEPS(+) & $-$ & $-$ & $-$ & $32.93$ & $38.58$ & $5.14$ & $29.26$ & $35.53$ & $5.68$ \\
& [Depth + PS](+)  & $-$ & $-$ & $-$ & $36.06$ & $41.83$ & $4.88$ & $32.29$ & $38.26$ & $5.33$ \\
\cmidrule{2-11}
& \netname (Ours) & $-$ & $-$ & $-$ & $\mathbf{38.83}$ & $\mathbf{44.91}$ & $\mathbf{4.37}$ & $\mathbf{33.82}$ & $\mathbf{39.92}$ & $\mathbf{4.82}$ \\
\bottomrule
\end{tabular}
\label{tab:combined_results}
\end{table*}

\subsection{Depth Decoder}
For depth estimation, we adopt the Monodepth2~\cite{godard2019digging} architecture, which has five convolutional layers with skip connections to the encoder. The depth decoder takes the predicted future multi-scale features and outputs depth maps $D_{t+\Delta}$ at different time steps $\Delta$. In addition to depth estimation, we use PoseNet to estimate the relative camera motion between image pairs. PoseNet consists of a ResNet-18 backbone followed by a four-layer convolutional network. During training, we enforce supervision through a photometric loss, which measures the pixel-wise difference between the original and reconstructed images using the predicted depth and pose.\looseness=-1

\subsection{Panoptic Decoder}
For panoptic segmentation, we implement a Panoptic-Deeplab-based decoder~\cite{cheng2020panoptic}. This bottom-up approach consists of two heads: one for semantic segmentation and the other for instance segmentation. The semantic segmentation head uses a fully convolutional network to predict a semantic label for each pixel. The instance segmentation head consists of two sub-heads. One sub-head predicts the center of each object, while the other assigns each pixel to the corresponding object center. A fusion module combines the predictions from both heads. For each instance, a majority voting mechanism is applied to assign the most frequent semantic label to the object, thus completing the panoptic segmentation task.

\begin{table*}
\caption{Comparison of \metricname results on the KITTI-360 dataset for different future frames and depth error thresholds. We report the average \metricname score and specific \metricname values for varying depth error thresholds (0.1, 0.25, and 0.5) at four future time steps: $t+0$, $t+1$, $t+3$, and $t+5$. Higher values mean better performance in panoptic-depth forecasting, with the proposed metric coherently capturing panoptic and geometric accuracy of future predictions.}
\centering
\footnotesize
\begin{tabular}{p{2.3cm}|P{0.5cm}P{0.5cm}P{0.5cm}P{0.5cm}|P{0.5cm}P{0.5cm}P{0.5cm}P{0.5cm}|P{0.5cm}P{0.5cm}P{0.5cm}P{0.5cm}|P{0.5cm}P{0.5cm}P{0.5cm}P{0.5cm}}
\toprule
Network & \multicolumn{4}{c|}{\metricname $t+0$} & \multicolumn{4}{c|}{\metricname $t+1$} & \multicolumn{4}{c|}{\metricname $t+3$} & \multicolumn{4}{c}{\metricname $t+5$} \\
\cmidrule{2-17}
 & \multicolumn{4}{c|}{$\Delta t = 0$} & \multicolumn{4}{c|}{Short term $\Delta t = 1$} & \multicolumn{4}{c|}{Short term $\Delta t = 3$} & \multicolumn{4}{c}{Mid term $\Delta t = 5$} \\
\cmidrule{2-17}
 & $avg$ & $0.1$ &$0.25$ & $0.5$ & $avg$ & $0.1$ &$0.25$ & $0.5$ & $avg$ & $0.1$ &$0.25$ & $0.5$ & $avg$ & $0.1$ &$0.25$ & $0.5$ \\
\midrule

CoDEPS(+) & $39.04$ & $38.44$ & $39.19$ & $39.49$ & $36.25$ & $25.67$ & $36.4$ & $36.69$ & $30.01$ & $29.67$ & $30.31$ & $30.55$  & $25.39$ & $24.93$& $25.52$& $25.73$\\

[Depth + PS](+) & $42.01$ & $41.28$ & $42.16$ & $42.61$ & $38.35$ & $37.66$ & $38.51$ & $38.89$ & $32.45$ & $31.85$ & $32.6$ & $32.89$  & $28.47$ & $27.96$& $28.6$& $28.8$ \\

\midrule
\netname (Ours) & $\mathbf{43.91}$ & $\mathbf{43.01}$ & $\mathbf{44.11}$& $\mathbf{44.63}$ & $\mathbf{41.03}$ & $\mathbf{40.26}$ & $\mathbf{41.21}$& $\mathbf{41.61}$ & $\mathbf{36.26}$ & $\mathbf{35.61}$& $\mathbf{36.42}$& $\mathbf{36.75}$ & $\mathbf{32.17}$& $\mathbf{31.56}$& $\mathbf{32.33}$& $\mathbf{32.61}$  \\
\bottomrule
\end{tabular}
\label{tab:metric}
\end{table*}

\subsection{Training Loss}
We train our \netname architecture using a combination of supervised and unsupervised losses to train both the depth estimation and panoptic segmentation decoders~\cite{voedisch23codeps}. For the depth decoder, we follow the standard unsupervised methodology based on photometric error~\cite{zhou2017unsupervised}. Given an image triplet $\{I_{t0}, I_{t1}, I_{t2}\}$, we predict the depth $D_{t1}$ and the camera motion $M_{t0 \rightarrow t1}$ and $M_{t1 \rightarrow t2}$. The depth photometric error loss is then calculated as a weighted sum of the reprojection loss $L_\text{pr}^d$ and image smoothness loss $L_\text{sm}^d$ as $L_\text{pe}^d = \lambda_\text{pr} L_\text{pr}^d + \lambda_\text{sm} L_\text{sm}^d$.

For the semantic segmentation decoder, we use a supervised bootstrapped cross-entropy loss with hard pixel mining $L_\text{bce}^\text{sem}$ as presented in Panoptic-Deeplab~\cite{cheng2020panoptic}. For the instance segmentation decoder, we use a mean squared error loss $L_\text{center}^\text{ins}$ for the instance center prediction and an L1 loss $L_\text{offset}^\text{ins}$ for the instance offset prediction. The total instance segmentation loss is computed as a weighted sum $L_\text{co}^\text{ins} = \lambda_\text{center} L_\text{center}^\text{ins} + \lambda_\text{offset} L_\text{offset}^\text{ins}$, where $\lambda_\text{offset}$ and $\lambda_\text{center}$ are $0.1$ and $10$. Finally, we compute the total training loss as the sum of all losses as $L_\text{total} = L_\text{pe}^d + L_\text{bce}^\text{sem} + L_\text{co}^\text{ins}$. This combined loss enables the model to learn both depth and panoptic segmentation jointly, ensuring that both tasks are effectively optimized during training.

\section{Experimental Evaluation}
In this section, we detail our training protocol and present comprehensive results on KITTI-360~\cite{liao2022kitti} and Cityscapes~\cite{cordts2016cityscapes} datasets, demonstrating the effectiveness of our approach on the three forecasting tasks.




\begin{table*}
\caption{Panoptic-forecasting results on the Cityscapes dataset. The table compares panoptic quality (PQ), segmentation quality (SQ), and recognition quality (RQ) for short-term ($\Delta t = 3$) and mid-term ($\Delta t = 9$) panoptic segmentation forecasts. All values are presented in $\%$.}
\centering
\footnotesize
\begin{tabular}{p{2.8cm}|P{0.35cm}P{0.35cm}P{0.35cm}|P{0.35cm}P{0.35cm}P{0.35cm}|P{0.35cm}P{0.35cm}P{0.35cm}|P{0.35cm}P{0.35cm}P{0.35cm}|P{0.35cm}P{0.35cm}P{0.35cm}|P{0.35cm}P{0.35cm}P{0.35cm}}
\toprule
& \multicolumn{9}{c|}{Short term $\Delta t = 3$} & \multicolumn{9}{c}{Mid term $\Delta t = 9$}
\\
\cmidrule{2-19}
& \multicolumn{3}{c|}{All} &\multicolumn{3}{c|}{Things} & \multicolumn{3}{c|}{Stuff} & \multicolumn{3}{c|}{All} &\multicolumn{3}{c|}{Things} & \multicolumn{3}{c}{Stuff}
\\
\cmidrule{2-19}
Network & PQ & SQ & RQ & PQ & SQ & RQ & PQ & SQ & RQ & PQ & SQ & RQ & PQ & SQ & RQ & PQ & SQ & RQ  \\
\midrule
Oracle $\dagger$ & $60.3$ & $81.5$&  $72.9$ & $51.1$ & $80.5$&  $63.5$& $67.0$ & $82.3$&  $79.7$& $60.3$ & $81.5$&  $72.9$& $51.1$ & $80.5$&  $63.5$& $67.0$ & $82.3$&  $79.7$\\
\midrule
Deeplab(Lastseen frame)& 
$32.7$ & $71.3$&  $42.7$ & $22.1$ & $68.4$&  $30.8$ & $40.4$ & $73.3$ & $51.4$ & $22.4$ & $68.5$&  $30.4$ & $10.7$ & $35.1$ & $80.5$ & $63.5$& $82.3$ & $79.7$\\
Flow \cite{graber2021panoptic}  & 
$41.4$ & $73.4$&  $53.4$ & $30.6$ & $70.6$&  $42.0$ & $49.3$ & $75.4$ & $61.8$ & $25.9$ & $69.5$&  $34.6$ & $13.4$ & $67.1$ & $19.3$ & $35.0 $& $71.3$ & $45.7$\\
F2MF \cite{vsaric2021dense}& $47.3$ & $75.1$ & $60.6$ & $-$ & $-$ & $-$ & $-$ & $-$ & $-$ & $33.1$ & $71.3$ & $43.3$ &  $-$ & $-$ & $-$ & $-$ & $-$ & $-$\\
IndRNN-Stack \cite{graber2021panoptic} & $49.0$ & $74.9$ & $63.3$ & $40.1$ & $72.5$ & $54.6$ & $55.5$ & $76.7$ & $69.5$ & $36.3$ & $71.3$ & $47.8$ & $25.9$ & $69.0$ & $36.2$ & $43.9$ & $72.9$ & $56.2$\\
DiffAttn-Fuse \cite{graber2022joint} & $50.2$ & $75.7$ & $64.3$ & $42.4$ & $74.2$ & $56.5$ & $55.9$ & $76.8$ & $70.0$ & $36.6$ & $71.4$ & $49.5$ & $28.6$ & $69.0$ & $40.1$ & $44.1$ & $73.2$ & $56.4$\\
\midrule
\netname (Ours) & $\mathbf{50.7}$ & $\mathbf{77.2}$ & $\mathbf{63.3}$ & $\mathbf{41.8}$ & $\mathbf{74.8}$ & $\mathbf{55.6}$ & $\mathbf{57.1}$ & $\mathbf{78.9}$ & $\mathbf{68.9}$ & $\mathbf{37.7}$ & $\mathbf{68.3}$ & $\mathbf{55.6}$ & $\mathbf{25.2}$ & $\mathbf{72.1}$ & $\mathbf{35.0}$ & $\mathbf{46.9}$ & $\mathbf{65.4}$ & $\mathbf{70.6}$\\
\bottomrule
\end{tabular}
\label{tab:psforecasting}
\end{table*}

\begin{table*}
\caption{Depth-forecasting results on KITTI-eigen. The table reports depth estimation metrics such as Absolute Relative Error (Abs Rel), RMSE, and threshold accuracy ($\delta<1.25$, $\delta<1.25^{2}$, $\delta<1.25^{3}$) across short-term ($\Delta t = 1$ and $\Delta t = 3$) and mid-term ($\Delta t = 5$) forecasting. The baselines only report depth forecasting for $\Delta t =5$}.
\centering
\footnotesize
\begin{tabular}{p{2.8cm}|P{0.5cm}P{0.5cm}P{0.5cm}P{0.5cm}P{0.5cm}|P{0.5cm}P{0.5cm}P{0.5cm}P{0.5cm}P{0.5cm}|P{0.5cm}P{0.5cm}P{0.5cm}P{0.5cm}P{0.5cm}}
\toprule
& \multicolumn{5}{c}{Short term $\Delta t = 1$} & \multicolumn{5}{c}{Short term $\Delta t = 3$} & \multicolumn{5}{c}{Mid term $\Delta t = 5$}
\\
\cmidrule{2-16}
Network & Abs Rel & RMSE & $\delta< 1.25$ & $\delta< 1.25^{2}$ & $\delta< 1.25^{3}$ & Abs Rel & RMSE & $\delta< 1.25$& $\delta< 1.25^{2}$ & $\delta< 1.25^{3}$ & Abs Rel & RMSE  & $\delta< 1.25$ & $\delta< 1.25^{2}$ & $\delta< 1.25^{3}$\\
  & $\downarrow$& $\downarrow$ & $\uparrow$ &$\uparrow$ & $\uparrow$& $\downarrow$& $\downarrow$ & $\uparrow$ &$\uparrow$ & $\uparrow$& $\downarrow$& $\downarrow$ & $\uparrow$ &$\uparrow$ & $\uparrow$ \\
\midrule
Oracle $\dagger$ & $0.098$ & $4.459$&  $0.900$ & $0.965$ & $0.983$&  $0.098$& $4.459$ & $0.900$&  $0.965$& $0.983$ & $0.098$&  $4.459$& $0.900$ & $0.965$&  $0.983$\\
\midrule

ForecastMonodepth2 \cite{boulahbal2022forecasting} & $-$ &$-$ &$-$ &$-$ &$-$ &$-$ &$-$ &$-$ &$-$ &$-$ &  $0.201$ & $6.166$ & $0.702$ & $0.897$ & $0.960$ \\

DepthEgomotion(+) \cite{boulahbal2022forecasting} & $-$ &$-$ &$-$ &$-$ &$-$ &$-$ &$-$ &$-$ &$-$ &$-$ &  $0.178$ & $6.196$ & $0.761$ & $0.914$ & $0.964$ \\

\midrule
\netname (Ours) &
$\mathbf{0.152}$ & $\mathbf{5.639}$ & $\mathbf{0.841}$ & $\mathbf{0.935}$ & $\mathbf{0.966}$ & $\mathbf{0.158}$ & $\mathbf{5.690}$ & $\mathbf{0.831}$ & $\mathbf{0.934}$ & $\mathbf{0.966}$ & $\mathbf{0.166}$ & $\mathbf{5.852}$ & $\mathbf{0.817}$ & $\mathbf{0.931}$ & $\mathbf{0.965}$ \\
\bottomrule
\end{tabular}
\label{tab:dforecasting}
\vspace{-0.2cm}
\end{table*}

\begin{figure*}
\setlength{\tabcolsep}{0.005cm} 
\begin{tabular}{P{0.3cm} P{5.70cm} P{5.70cm} P{5.70cm}}
  & \scriptsize $\Delta t = 1$ & \scriptsize $\Delta t = 3$ & \scriptsize $\Delta t = 5$ \\
    \rotatebox{90}{\tiny Image} &
    \includegraphics[width=0.49\linewidth]{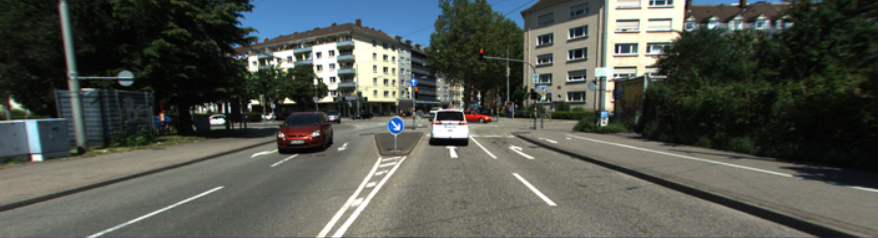} &
    \includegraphics[width=0.49\linewidth]{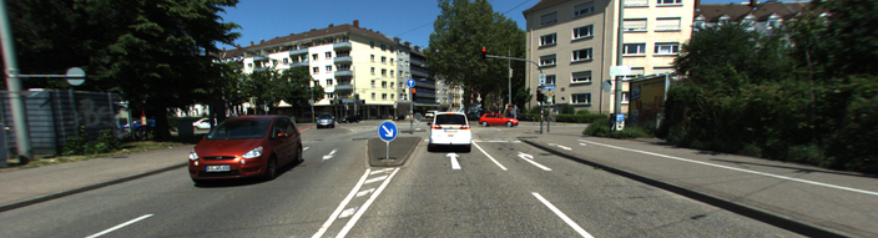} &
    \includegraphics[width=0.49\linewidth]{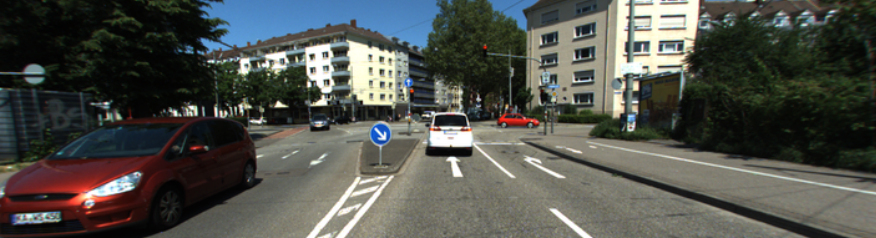} \\
    \rotatebox{90}{\tiny GT} &
    \includegraphics[width=0.49\linewidth]{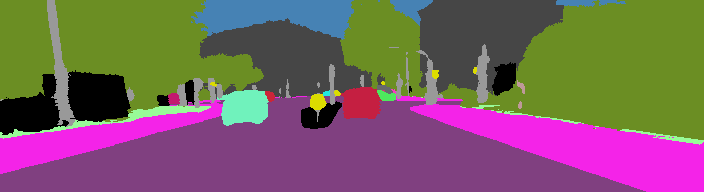}%
    \includegraphics[width=0.49\linewidth]{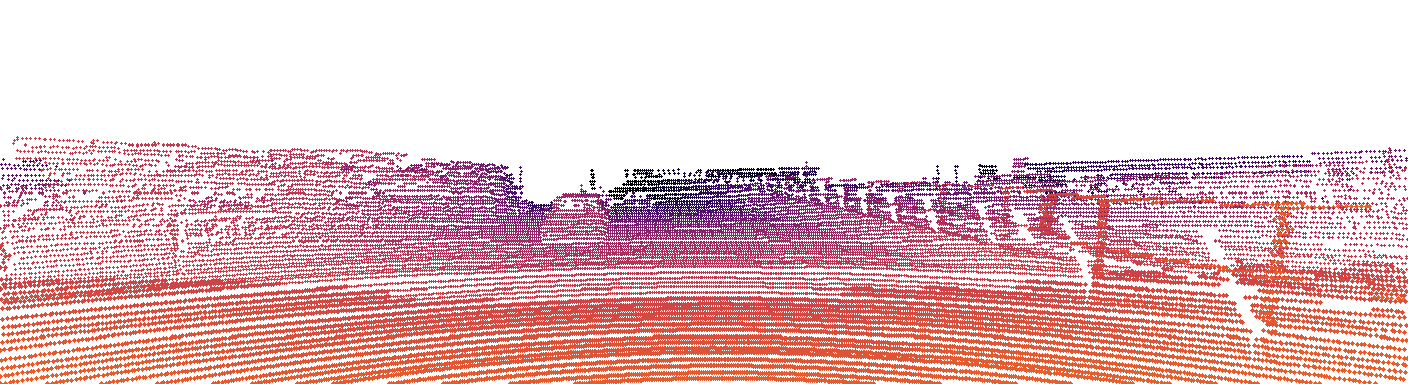} &
    \includegraphics[width=0.49\linewidth]{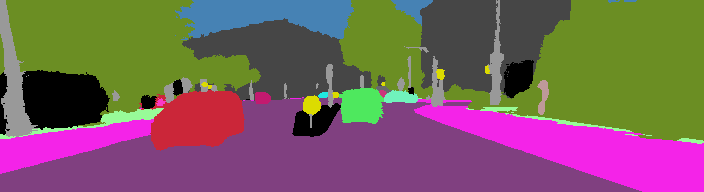}%
    \includegraphics[width=0.49\linewidth]{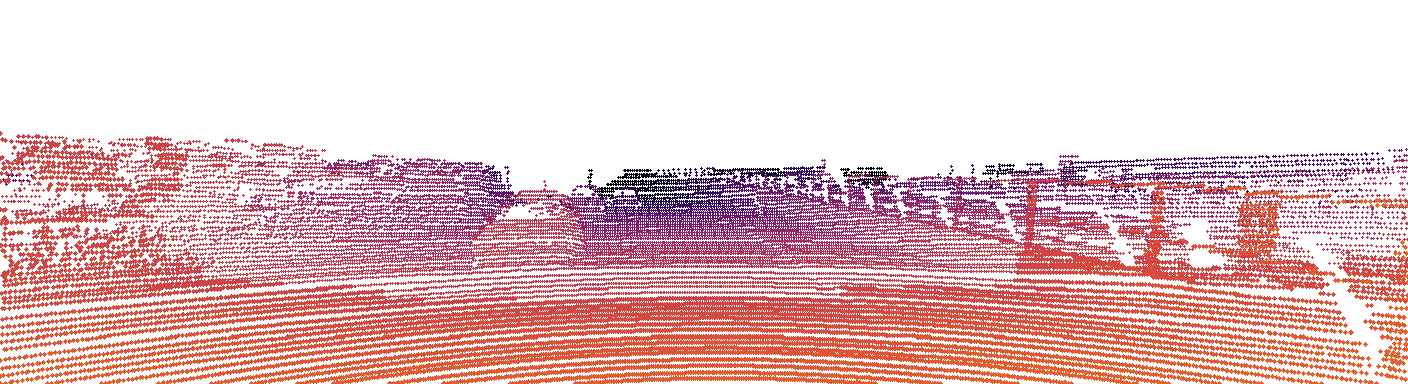} &
    \includegraphics[width=0.49\linewidth]{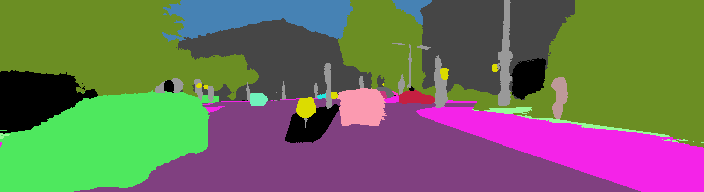}%
    \includegraphics[width=0.49\linewidth]{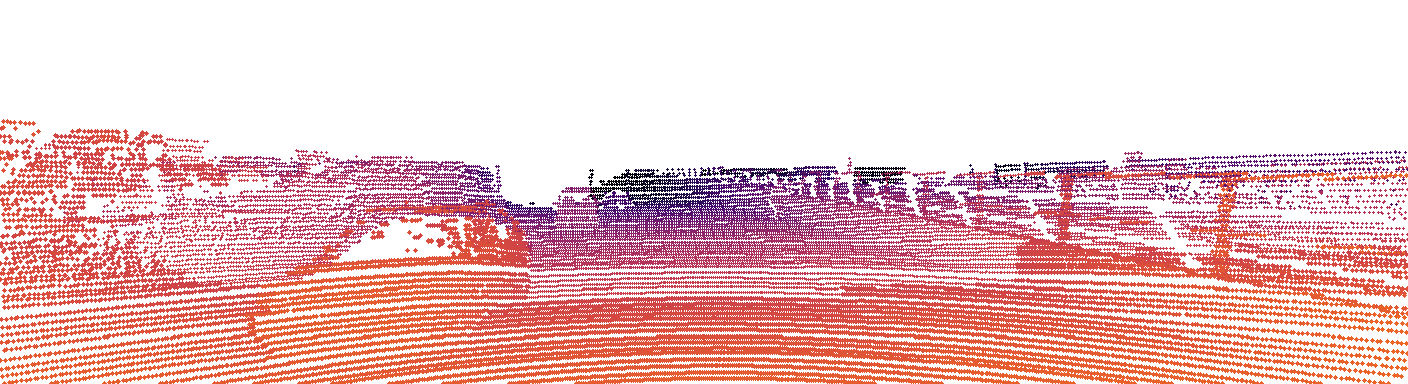} \\
    \rotatebox{90}{\tiny CoDEPS(+)} &
    \includegraphics[width=0.49\linewidth]{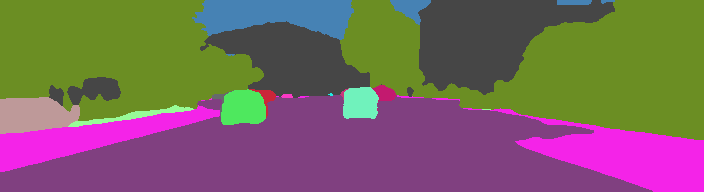}%
    \includegraphics[width=0.49\linewidth]{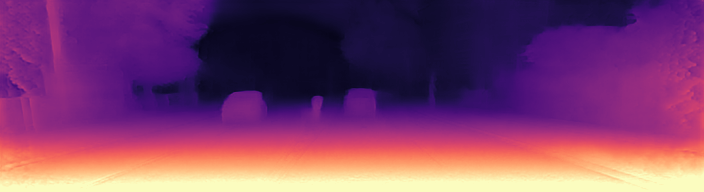} &
    \includegraphics[width=0.49\linewidth]{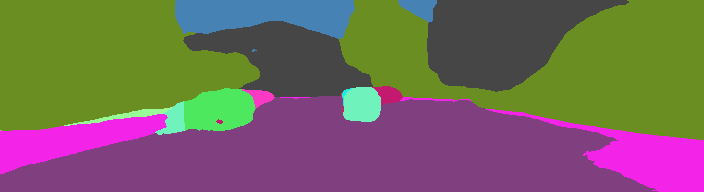}%
    \includegraphics[width=0.49\linewidth]{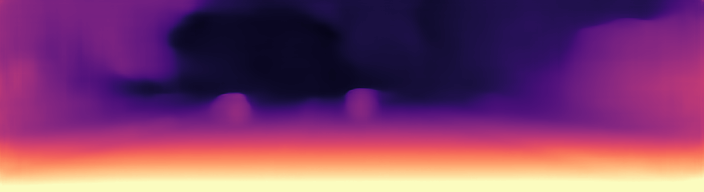} &
    \includegraphics[width=0.49\linewidth]{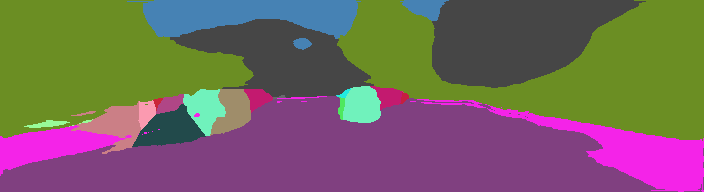}%
    \includegraphics[width=0.49\linewidth]{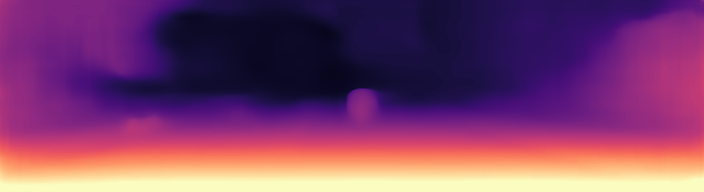} \\
    \rotatebox{90}{\tiny \netname} &
    \includegraphics[width=0.49\linewidth]{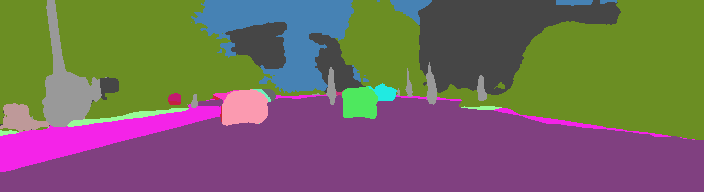}%
    \includegraphics[width=0.49\linewidth]{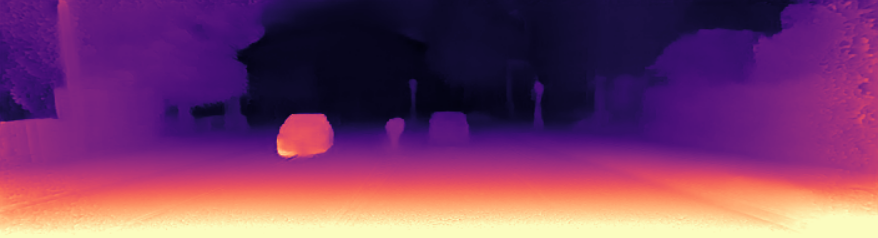} &
    \includegraphics[width=0.49\linewidth]{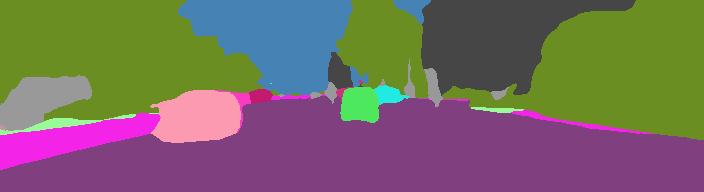}%
    \includegraphics[width=0.49\linewidth]{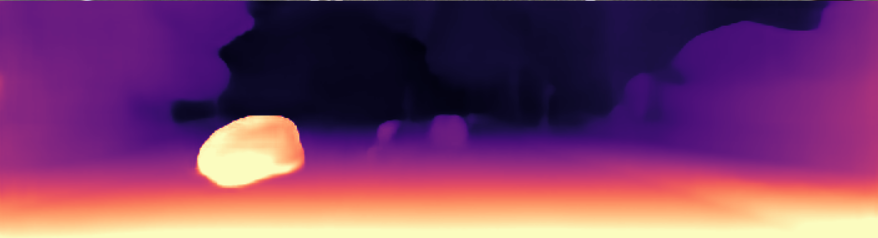} &
    \includegraphics[width=0.49\linewidth]{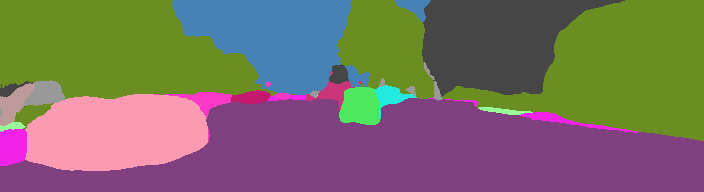}%
    \includegraphics[width=0.49\linewidth]{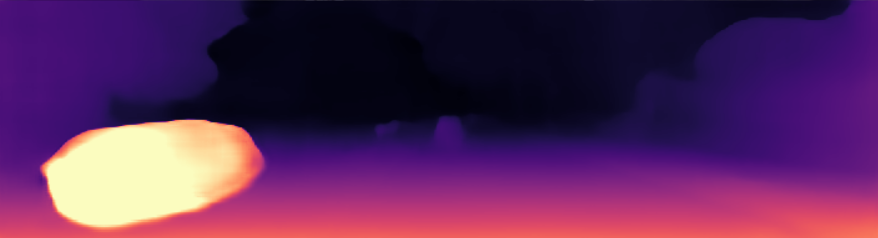} \\
\end{tabular}
\caption{Qualitative comparison of predictions from our proposed \netname model with the second-best baseline CoDEPS(+) on the KITTI-360 dataset. We show the camera image corresponding to the future frame at $t + \Delta$ and the panoptic-depth ground truth (GT). We observe that our model accurately forecasts panoptic-depth predictions, capturing scene details such as poles, even when a car is exiting the frame.}
\label{fig:qualitative}
\vspace{-0.3cm}
\end{figure*}

\subsection{Training Protocol}
For KITTI-360, we use images of resolution $192 \times 704$~pixels and retrieve depth maps from LiDAR point clouds. Cityscapes-DVPS~\cite{qiao2021vip} provides re-computed depth maps from the Cityscapes dataset consisting of images with a resolution of $1024 \times 2048$~pixels, and we generate depth maps from disparity images. For both datasets, we use the semantic and instance segmentation annotations and generate additional center heatmaps and $(x, y)$ offset maps. We initialize the encoder with swin-tiny pretrained weights~\cite{liu2021swin} and the decoders with pretrained weights from CoDEPS~\cite{voedisch23codeps}. We optimize our model using the Adam optimizer with a learning rate $lr =0.0001$. 

\subsection{Benchmarking Results}

We compare our proposed \netname architecture with two baseline models on the KITTI-360 and Cityscapes-DVPS datasets. As shown in Tab.~\ref{tab:combined_results}, PDCast consistently outperforms both baselines across all time steps. For short-term predictions ($\Delta t = 1$), PDCast achieves a \metricname score of 41.03 and an RMSE of 4.04, outperforming both baselines by a substantial margin. The improvement continues for mid-term predictions ($\Delta t = 5$), where \netname achieves a \metricname score of $32.17$, outperforming the baselines by up to $6.78\%$. Notably, \netname exhibits better panoptic quality (PQ) and root mean square error (RMSE), indicating more accurate joint panoptic and depth forecasting. Our proposed \metricname metric effectively measures panoptic-depth forecasting by evaluating both panoptic segmentation quality and depth accuracy. As shown in Tab.~\ref{tab:metric}, the metric assesses the performance across different future time steps and depth error thresholds (0.1, 0.25, 0.5). Our method performs better than the baselines at all time steps, especially for mid-term predictions ($t+5$), where it achieves a \metricname score of 32.17 compared to 25.39 for CoDEPS(+). The \metricname metric is able to assess forecasting accuracy with a focus on both semantics and depth, making it fitting for evaluating joint panoptic and depth predictions.

The results for the panoptic forecasting task on the Cityscapes dataset presented in Tab.~\ref{tab:psforecasting} show that our method achieves competitive results, particularly in the 'thing' category for both short-term ($\Delta t = 3$) and mid-term ($\Delta t = 9$) predictions. Our model also presents strong performance in the 'stuff' category and consistently ranks among the top performers, demonstrating its robustness across various classes. Furthermore, our method achieves state-of-the-art performance in depth forecasting on the KITTI-eigen benchmark, as shown in Table~\ref{tab:dforecasting}. Across all forecasting horizons, including short-term ($\Delta t = 1$ and $\Delta t = 3$) and mid-term ($\Delta t = 5$), our approach outperforms existing methods. Specifically, our model demonstrates lower Absolute Relative Error (Abs Rel) and RMSE while maintaining competitive threshold accuracy ($\delta < 1.25$, $\delta < 1.25^{2}$, $\delta < 1.25^{3}$).

\subsection{Qualitative Results}
We qualitatively compare the performance of \netname with the second-best baseline CoDEPS(+) on the KITTI-360 dataset, shown in Fig.~\ref{fig:qualitative}. Across the different future time steps, we observe that \netname consistently yields more detailed panoptic-depth predictions. For example, \netname accurately captures scene elements such as poles and vehicles, even when a car is exiting the frame at $\Delta t = 5$. With PDcast, the semantic segmentation boundaries are more defined, and the depth predictions accurately align with the scene geometry. 

In contrast, we observe that CoDEPS(+) generates less defined scene elements as $\Delta t$ increases. The panoptic and depth predictions are particularly inaccurate at $\Delta t = 5$, where CoDEPS(+) ignores the car exiting the bottom left side of the frame. Additionally, the instance segmentation is notably inaccurate as the overall scene structure lacks the detail that is preserved by PDcast. This highlights our network's ability to retain finer details of future scene geometry and semantics.
\section{Conclusion}
\label{sec:conclusions}

In this paper, we introduced the novel panoptic-depth forecasting task, which jointly predicts panoptic segmentation and depth maps of unobserved future frames, from moncular camera images as input. We proposed the \metricname metric to coherently evaluate this new task by combining panoptic quality and depth accuracy across varying time horizons and error thresholds. More importantly, we proposed the novel \netname architecture that achieves state-of-the-art performance across multiple datasets and forecasting tasks, consistently outperforming baseline models. 

The results demonstrate the benefits of joint panoptic-depth forecasting as our \netname model exceeds the performance of specialized individual panoptic forecasting and depth forecasting methods, showcasing its versatility in holistic scene understanding. Furthermore, we made the code and pretrained models publicly available. Our contributions provide a solid foundation for future research in holistic forecasting frameworks for autonomous systems.

\footnotesize
\bibliographystyle{IEEEtran}
\bibliography{references.bib}

\end{document}